\documentclass[a4paper]{svproc}

\usepackage{url}

\usepackage{graphics} %
\graphicspath{{images/}{img/}{generated-images/}{results/}}
\usepackage{epsfig} %
\usepackage{amsmath} %
\usepackage{amssymb}  %
\usepackage{cite}
\usepackage{gensymb}
\usepackage{subfig}
\usepackage{todonotes}
\usepackage{multirow}
\usepackage[export]{adjustbox}
\usepackage{colortbl}
\usepackage[nolist]{acronym}
\usepackage{hyperref}

\definecolor{todo-red}{RGB}{200,12,12}
\definecolor{PRGreen}{RGB}{54,148,90}
\definecolor{RayOrange}{RGB}{204,102,11}
\definecolor{OccYellow}{RGB}{255,251,212}
\definecolor{FreeGreen}{RGB}{211,233,232}
\definecolor{UnkBlue}{RGB}{218,205,221}
\definecolor{ConsRed}{RGB}{170,97,97}

\newcommand{\reffig}[1]{Fig.~\ref{#1}}

\newcommand{\refsec}[1]{Section~\ref{#1}}
\newcommand{\qmarks}[1]{``#1"}
\newcommand{\norm}[1]{\left\lVert#1\right\rVert}

\newcommand{\ffs}{\mathcal{V}_{\text{free}}} %
\newcommand{\kfs}{\bar{\mathcal{V}}_{\text{free}}} %
\newcommand{\bkfs}{\partial \kfs} %
\newcommand{\vfov}{\mathcal{V}_{\text{fov}}(\mathtt{T}_{W,R})}
\newcommand{\vfovof}[1]{\mathcal{V}_{\text{fov}}(\mathtt{T}_{W,R_{#1}})}
\newcommand{\vfovk}{\vfovof{k}}
\newcommand{\pose}[2]{\mathtt{T}_{#1,#2}}
\newcommand{\relpose}{\pose{R_{k-1}}{R_k}}
\newcommand{\posecomm}{\posest{R_{k-1}}{R_k}}
\newcommand{\posest}[2]{\bar{\mathtt{T}}_{#1,#2}}
\newcommand{\posiest}[2]{\bar{\mathbf{t}}_{#1,#2}}
\newcommand{\noise}[1]{\eta_{#1}}

\newcommand{\dfov}{d_\text{FOV}}
\newcommand{\dexp}{d_\text{exp}}
\newcommand{\dmax}{d_\text{max}}
\newcommand{\consrad}{\mathcal{R}}
\newcommand{\lcd}{d_\text{pr}}

\definecolor{somegray}{rgb}{0.5, 0.5, 0.5}
\newcommand{\darkgrayed}[1]{\textcolor{somegray}{#1}}

\begin{document}

\begin{acronym}
	\acro{TSDF}{Truncated Signed Distance Field}
	\acro{NBV}{Next-best-view}
	\acro{FOV}{Field of View}
	\acro{SLAM}{Simultaneous Localisation and Mapping}
\end{acronym}

\mainmatter              %
\title{Exploration Without Global Consistency \\ Using Local Volume Consolidation}
\titlerunning{Exploration without Global Consistency}  %
\author{Titus Cieslewski \and Andreas Ziegler \and Davide Scaramuzza}
\authorrunning{Titus Cieslewski et al.} %
\tocauthor{Titus Cieslewski, Andreas Ziegler, and Davide Scaramuzza}
\institute{Robotics and Perception Group, Depts. of Informatics and Neuroinformatics, University of Zurich and ETH Zurich \\
\url{http://rpg.ifi.uzh.ch}
}

\maketitle              %

\vspace{-6cm}
\noindent \darkgrayed{This paper has been accepted for publication at the International Symposium on Robotics Research (ISRR), Hanoi, 2019.}
\vspace{5cm}

\begin{abstract}
In exploration, the goal is to build a map of an unknown environment. 
Most state-of-the-art approaches use map representations that require drift-free state estimates to function properly. 
Real-world state estimators, however, exhibit drift.
In this paper, we present a 2D map representation for exploration that is robust to drift. 
Rather than a global map, it uses local metric volumes connected by relative pose estimates.
This pose-graph does not need to be globally consistent.
Overlaps between the volumes are resolved locally, rather than on the faulty estimate of space.
We demonstrate our representation with a frontier-based exploration approach, evaluate it under different conditions and compare it with a commonly-used grid-based representation. 
We show that, at the cost of longer exploration time, using the proposed representation allows full coverage of space even for very large drift in the state estimate, contrary to the grid-based representation.
The system is validated in a real world experiment and we discuss its extension to 3D.
\\
\\
{\bf Video:} A video is available at \url{https://youtu.be/s4Xnet_h4ss}
\end{abstract}

\vspace{-5mm}

\begin{figure}
\centering
\subfloat[Global, grid-based]{\includegraphics[width=0.48\linewidth]{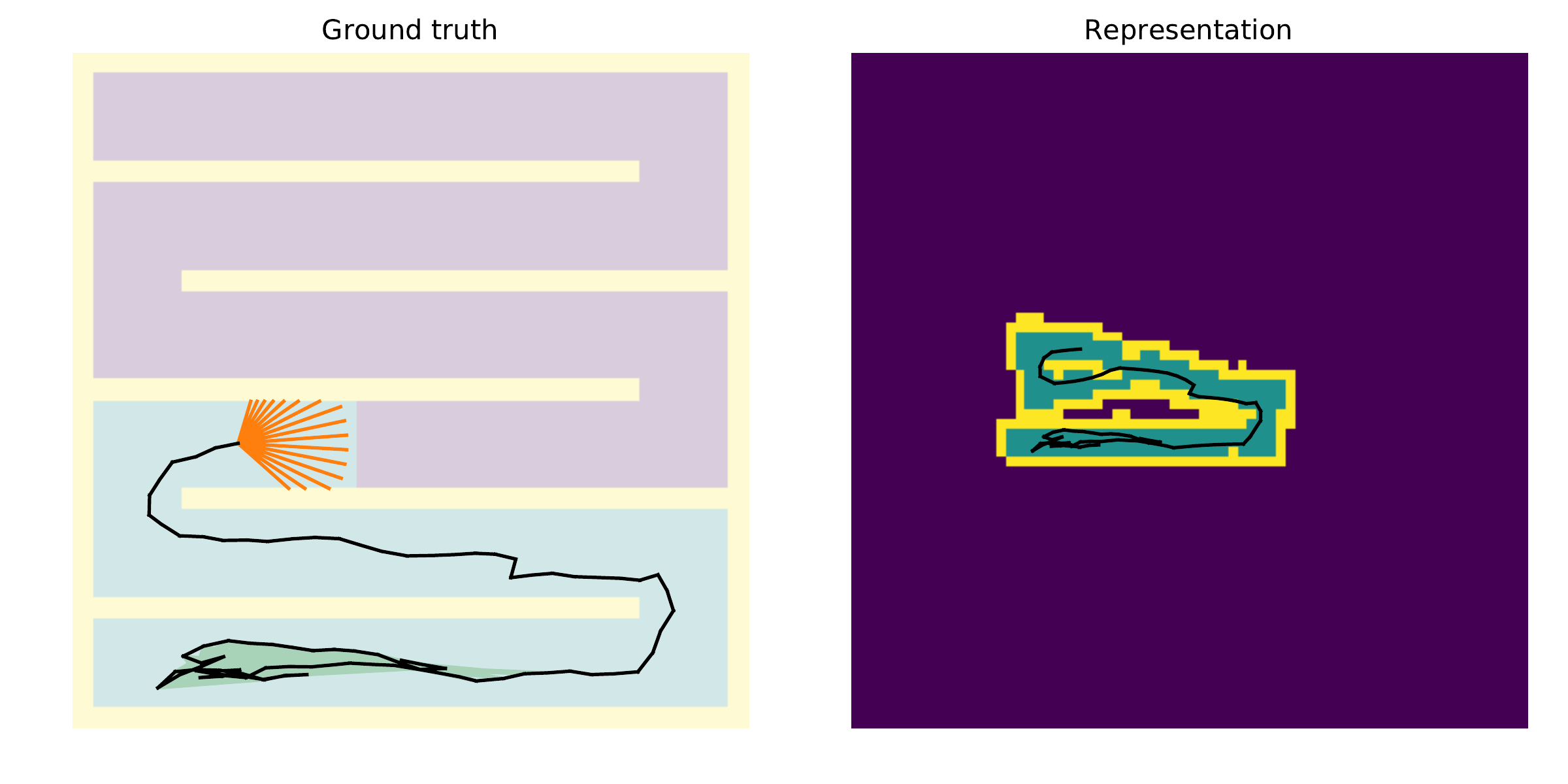}}
\hfil
\subfloat[Ours]{\includegraphics[width=0.48\linewidth]{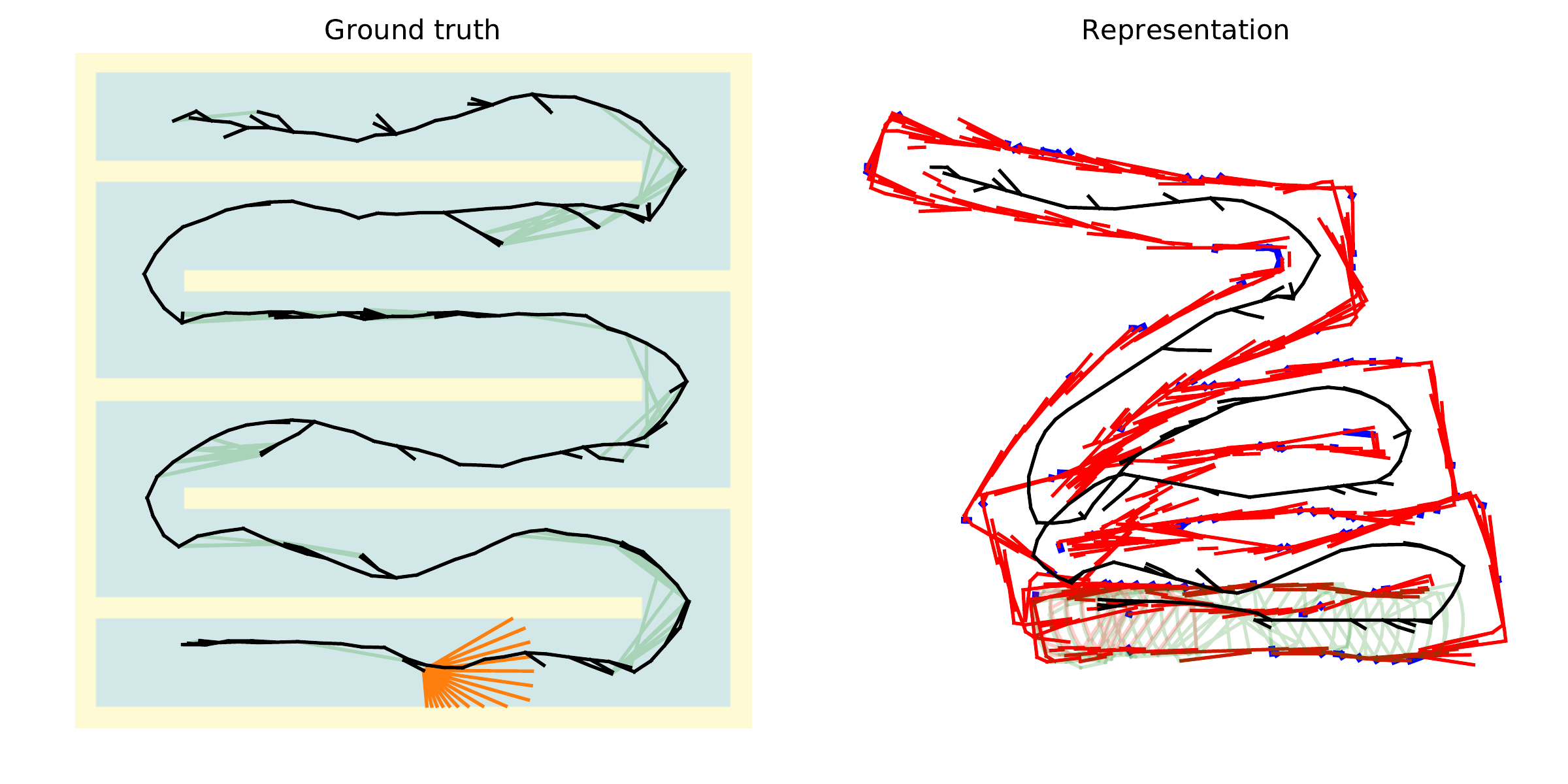}}
\vspace{-3mm}
\caption{
In a maze-like environment where loop closures cannot be established to account for pose estimate drift, grid-based representations build an inconsistent map, which can lead to premature termination of exploration (a).
The purple region in the ground truth of explored space, in the leftmost image, indicates unexplored space.
Our method, in contrast, only resolves overlaps between volumes locally, which allows to fully explore the environment (b).
See \reffig{fig:repr} for a detailed legend.}
\label{fig:eyecat}
\end{figure}

\section{Introduction}\label{sec:introduction}

Exploration is a fundamental task in autonomous robotics. 
The goal is to map an unknown environment~\cite{Yamauchi97icra}, for example to establish how a robot can navigate through it.
Exploration can be used in various scenarios from search and rescue, oil and gas exploration, 3D reconstruction to inspection tasks. 
Different applications have different requirements. 
In search and rescue, for instance, it is important to find survivors rapidly. 
Other requirements could be that the generated map has a high accuracy or that the path travelled by the robot is as short as possible.
In exploration, a map representation is used to describe the space that the robot perceived so far.
Based on this description, a planning algorithm decides where to move next. 
In this work, we focus on the map representation.

Most map representations currently used for exploration assume perfect state estimates. 
However, on-board state estimators, such as visual-inertial odometry, are prone to drift.
Since these are the estimators most likely to be used in unknown environments in the real world, these map representations can in practice lead to incorrect maps, as illustrated in \reffig{fig:eyecat}. 

To some extent, drift can be removed when a robot revisits a place and a correction of the trajectory estimate can be performed with a loop closure. 
However, as \reffig{fig:eyecat} illustrates, not all drift error can be removed in this way.
Furthermore, such a correction causes heavy computational load in typical grid-based maps, as every depth measurement ray needs to be re-cast with the new trajectory estimate \cite{Wurm10icraw}.
Sub-mapping approaches can mitigate this computational cost~\cite{Bosse03icra},~\cite{Schmuck16ifac},~\cite{Millane17arxiv}.

\subsection{Contributions}

In this paper, we propose a map representation for exploration that is robust to state estimate drift.
We build our approach using ideas from~\cite{Howard06ieee}, where free space is represented with local polygons connected by relative poses.
Our contributions are as follows:
\begin{itemize}
\item We apply the ideas from~\cite{Howard06ieee} to exploration by adding semantics to the local polygons, which incorporate information from other nearby polygons.
Nearby polygons are determined based on vicinity in a pose graph, as opposed to vicinity in an estimate of space { that is faulty due to drift}.
\item We show that with this, global map consistency is not required for the robot to know when exploration is complete.
With this, global map optimization is not needed, which is particularly interesting for multi-robot exploration.
\item We show how to apply our representation to an exploration algorithm that previously used a grid-based representation.
Local navigation is performed in local polygons, while navigation between remote locations in the map is performed using teach-and-repeat~\cite{Es15crv}.
\item Finally, we simulate our proposed map representation under different conditions, compare it to grid-based representations, and validate it in the real world. 
As we show, our method comes at a cost of longer exploration times, but given reliable place recognition, is able to deal with large drift in cases where grid-based representations fail.
\end{itemize}

\section{Related Work}

While a map representation is used to represent the space that the robot perceived so far, a planning algorithm has to plan where the robot should move next. 
In the following, we review these two aspects of exploration separately.

\subsection{Map Representations}

Exploration can be formalized as the problem of discovering all free space within a bounded volume.
Thus, the map representation needs to represent the spatial information necessary to achieve this.
Map representations are commonly divided into three categories~\cite{Wallgrun10book}:
metric representations, topological representations and hybrid, topometric representations.

{\bf Metric Representations}
Metric representations express locations as coordinates in a global reference frame.
The predominant type of metric representations used in exploration are grid-based representations.
In a grid-based representation, the volume is quantized into voxels, arranged in a grid. 
The voxels are then assigned appropriate labels, such as \qmarks{free}, \qmarks{unknown} or \qmarks{obstacle}~\cite{Yamauchi97icra}.
Voxels typically carry further information, such as occupancy probabilities~\cite{Moravec85icra, Wurm10icraw} or a signed distance to the closest obstacle~\cite{Izadi11siggraph, Oleynikova16rssw}.
In two dimensions, an alternative way of representing known free space is to represent it with a polygon~\cite{Gonzalez02ijrr, Caccavale18iros}.
Here, the inside of the polygon represents free space, while its boundaries can either represent obstacles or the interface to unknown space.
Because the location of all entities in fully metric representations are expressed in a global reference frame, they are highly reliant on accurate pose estimates:
wrong pose estimates would lead to depth measurements being inconsistent with the true scene geometry, and with each other.
As a consequence, the robot could wrongly believe occupied space to be free or inversely, free space to be occupied, which could have disastrous consequences for navigation.

{\bf Topological Representations}
Unlike metric representations, topological representations do not express locations as coordinates in a reference frame.
Instead, locations are represented as vertices in a graph.
Relationships between locations, such as adjacency, are expressed as edges.
In order to be navigable, these edges should carry enough information to allow a robot to move between adjacent vertices.
Vision-based navigation in such maps has been demonstrated using Visual Teach and Repeat~\cite{Furgale10jfr,Es15crv}.
While topological maps can also be derived from volumetric maps~\cite{Wallgrun10book, Bloechliger17arxiv}, we are not aware of any work that uses purely topological maps to directly represent volumetric information.
Pure topological maps are thus rarely used for exploration.
A notable exception to this are~\cite{Bosse03icra, Akdeniz15icra}.
Rather than explicitly keeping track of free space, their robots build a pose graph similar to~\cite{Es15crv}.
At every vertex of this graph, the robot adds to the graph a finite amount of adjacent candidate locations where the robot could move to.
Subsequently, these candidate locations are all visited and the process is repeated recursively unless a candidate turns out to correspond to a previously visited location.
However, candidate robot locations are only a very crude representation of free space, which makes it hard to estimate the actual coverage.

{\bf Hybrid Representations}
Hybrid, or {\it topometric}, representations are a combination of metric and topological representations, aimed at combining their advantages.
Strictly speaking, the aforementioned pose graphs were already topometric, since they contained metric positions and transformations.
Nonetheless, in this section we focus on representations which augment topological graphs with volumetric information.
An early topometric representation has been proposed in \cite{Howard06ieee}.
This representation consists of a set of polygons, each representing the free space measured around a specific pose of the robot.
The locations of these polygons are expressed relative to each other.
However,~\cite{Howard06ieee} does not consolidate the information of neighbouring polygons in a way that would be helpful for exploration.
As a consequence, in~\cite{Howard06ijrr2}, where~\cite{Howard06ieee} is used in an exploration system, the authors fall back on local occupancy grids for exploration.
In contrast, we extend~\cite{Howard06ieee} with a consolidation procedure that allows its direct application to exploration.
Furthermore, while~\cite{Howard06ijrr2} ends up building a globally consistent map, we show that global consistency is not required for exploration.
The use of local occupancy grid {\it submaps} is also proposed in~\cite{Schmuck16ifac,Millane17arxiv}, where the authors aim at producing a globally consistent map of the environment just like~\cite{Howard06ijrr2}.
Achieving global consistency requires map optimization, which in turn leads to costly re-calculation of the occupancy grid submaps, even if it can sometimes be avoided for small loop closure corrections~\cite{Schmuck16ifac}.
We show that all of this can be avoided since global consistency is not necessary for exploration.

\subsection{Path Planning for Exploration}

In order to understand how to build a good representation, we need to understand how it is used. 
In path planning for exploration, after every observation the robot has to decide where to move next. 
In order to achieve fast exploration, the robot should perceive unknown space as quickly as possible. 
There are two main planning approaches in the literature: \ac{NBV} planning and frontier-based planning.

{\bf \acf{NBV} Planning}
\ac{NBV} planners try to choose the next position in such a way that the perception is optimal according to a {\it utility} function. 
This problem has also been studied in the computer vision community~\cite{Connolly85icra}. 
Most \ac{NBV} planners determine the next best view by sampling candidate views, predicting their utility~\cite{Gonzalez02ijrr, Papachristos17icra}, and picking the view with the highest utility.
The advantage is that the utility function can be adapted to contain arbitrary terms and constraints, such as the motion model of the robot, or terms describing map accuracy. 
The disadvantage is that finding the view with optimal utility is usually intractable, hence a sampling method has to be applied.
Furthermore, and somewhat as a consequence, NBV planning commonly requires heavy computational load.

{\bf Frontier-Based Exploration Planning}
Frontier-based exploration planners follow a simple and efficient scheme. 
The strategy is to navigate to the closest boundary between known free and unknown space.
These boundaries are referred to as {\it frontiers}. 
Frontier-based exploration methods assume that navigating to a frontier will result in the exploration of new space. 
The original idea was introduced in~\cite{Yamauchi97icra}. 
Unlike \ac{NBV} planners, frontier-based planners do not natively support arbitrary terms or constraints.
However, it was shown that they cover the space faster than their \ac{NBV} counterparts~\cite{Holz10isr, Cieslewski17iros}.
In this work, we will evaluate our representation using a modified version of the frontier-based exploration method presented in~\cite{Cieslewski17iros}.
Frontiers are explicitly expressed in our method, so it is natural to adopt frontier-based exploration.

\section{Problem Statement}\label{sec:ps}

The goal of exploration is to build a map of an unknown environment. 
Consider a bounded region of space $\mathcal{V} \in \mathbb{R}^2$ that represents the unknown environment that we would like to explore. 
$\mathcal{V}$ consists of free and occupied space $\mathcal{V} = \mathcal{V}_{\text{free}} \cup \mathcal{V}_{\text{occ}}$. 
With a robot that can measure the free space around itself, we can explore $\mathcal{V}_{\text{free}}$.
We denote a robot pose in the world frame $W$ as $\pose{W}{R}$ with orientation $\mathtt{R}_{W,R}$ and translation $t_{W,R}$.
The \acf{FOV} of the robot at pose $\mathtt{T}_{W,R}$ is denoted as $\vfov \subset \mathcal{V}_{\text{free}}$.
While a robot is moving on a trajectory $\mathtt{T}_{W,R}(t)$, it will measure $\mathcal{V}_{\text{fov}}$ at consecutive sampling times $t_0, t_1, \ldots$.
We denote the corresponding poses as $\mathtt{T}_{W,R_0}, \mathtt{T}_{W,R_1}, \ldots$. 
The space explored up to time $t_k$ ({\it known free space}) is the union of all the \acp{FOV} measured so far:
\begin{equation}\label{eq:fovunion}
  \kfs (t_k) = \displaystyle \bigcup_k \mathcal{V}_{\text{fov}}(\mathtt{T}_{W,R_k}).
\end{equation}
We consider two goals for our representation:
1) represent the robots' estimate of $\kfs$.
2) Allow to determine when exploration is complete, that is, when the whole free space is covered:
\begin{equation}\label{coverage}
\kfs = \mathcal{V}_{\text{free}}.
\end{equation}
Without knowing the ground truth $\mathcal{V}_{\text{free}}$, this condition can be established given sufficient knowledge about the boundary of $\kfs$, $\bkfs$.
Generally, $\bkfs$ either consists of interfaces to occupied space or interfaces to unknown free space.
As can be shown by contradiction, \eqref{coverage} holds if and only if $\bkfs$ consists only of interfaces to occupied space\footnote{
Trivial exceptions, such as disconnected free space, are omitted for brevity.} (\qmarks{no frontiers left}).
Hence, the second goal can be reached by correctly representing $\bkfs$.

\section{Proposed Representation}

\begin{figure}
    \centering
    \vspace{-5mm}
    \includegraphics[width=.8\columnwidth]{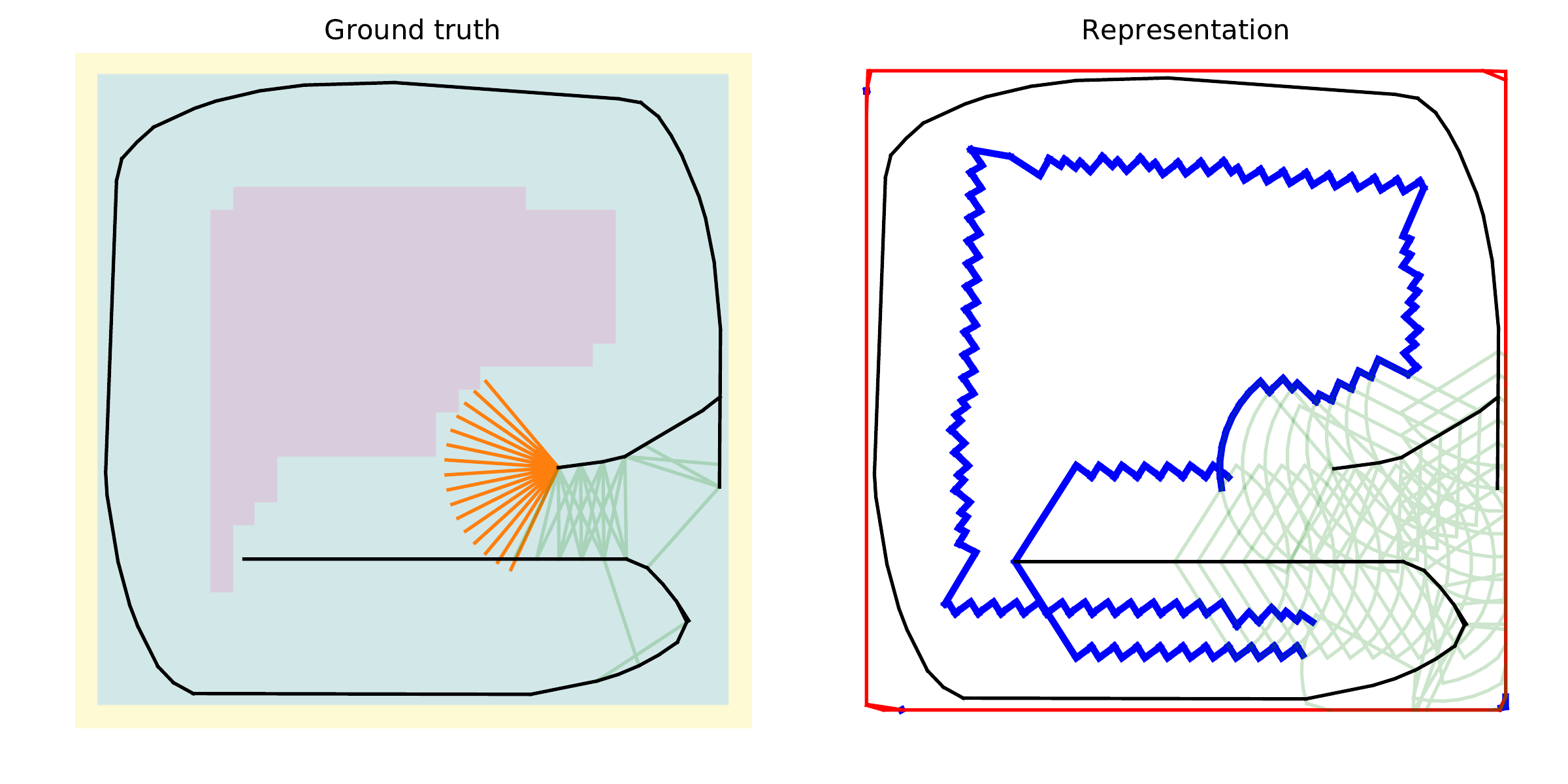}
    \caption{Our representation, visualized for a simulation without drift in the pose estimate.
    Left image (ground truth): black line: robot trajectory, {\color{PRGreen} green lines: place recognitions}, {\color{RayOrange} orange lines: depth sensor rays} at current pose; in the background, \colorbox{OccYellow}{yellow squares indicate true occupied space}, \colorbox{FreeGreen}{green squares true known free space} and \colorbox{UnkBlue}{blue squares true unknown free space}.
    Right image (our representation): black line: pose graph without loop closures, {\color{red}red lines: known obstacles}, {\color{blue} blue lines: frontiers}, {\color{PRGreen}green lines: local volumes in consolidation scope} (see \refsec{sec:consolidation}).
    Note how frontiers may overlap, as they are not resolved globally but only within the consolidation scope.}
    \label{fig:repr}
    \vspace{-10mm}
\end{figure}

As in~\cite{Howard06ieee}, the proposed map representation is based on polygons.
More specifically, local polygons individually represent the \ac{FOV} of every pose, $\vfov$.
The inside of a polygon represents known free space $\kfs$ while its edges represent the boundary $\bkfs$.
Unlike~\cite{Howard06ieee}, $\bkfs$ is labeled, which allows us to reach the second goal of the problem statement.
The labels are {\it obstacle} for parts of the boundary that are common with occupied space, {\it frontier} for parts adjacent to unknown space, and {\it free} for parts that are considered to lie within another polygon.
See \reffig{fig:repr} for a visualization of our representation.

\subsection{Local Volumes from Depth Measurements}\label{sec:construction}

\begin{figure}
    \vspace{2mm}
  \center
  \includegraphics[angle=90,width=0.40\textwidth]{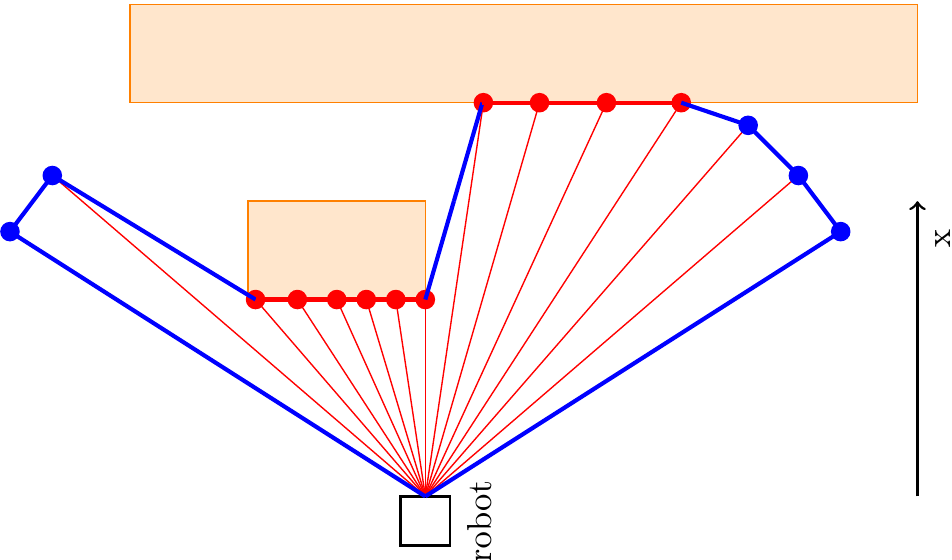}
  \caption{A local volume obtained from one measurement.
    The red lines starting at the robot position represent the depth measurement rays.
    \textcolor{blue}{Blue points represent samples at maximum sensor range} and \textcolor{red}{red points represent samples that hit an obstacle}.
    \textcolor{blue}{Blue edges are the frontier edges} and \textcolor{red}{red edges obstacle edges}.}
  \label{fig:fov}
  \vspace{-5mm}
\end{figure}

We assume a depth sensor as source to build the polygon of the robot's \acf{FOV}.
Each measurement consists of a set of simultaneously acquired depth samples.
In our simulations, these samples are modelled as equally distributed within the \ac{FOV}, as shown in \reffig{fig:fov}.
We build the polygon from these samples by taking the position of adjacent samples and the robot's position as vertices and connecting them with edges.
Some samples correspond to measured depths while others correspond to samples where the closest obstacle is beyond the sensor range.
If two adjacent depth samples are not beyond range, and have values with a difference below a threshold $\delta$, the edge connecting them is considered an obstacle edge.
In all other cases, the edge is considered a frontier edge.
The threshold $\delta$ is intended to account for occlusions (see \reffig{fig:fov}).
Sometimes, these cannot be distinguished from obstacle surfaces that are nearly parallel to the measurement rays~\cite{Gonzalez02ijrr}.
These incorrectly labeled frontiers can be resolved by approaching that obstacle from another angle.
Furthermore, the two edges adjacent to the robot position are also considered frontiers.

\subsection{Pose Graph Representation}\label{sec:posegraph}

In our representation, we store separate local volumes for each set of depth measurement samples, which are built as described in the previous section.
These polygons are referenced in the vertices of a {\it pose graph}.
Each vertex of this graph represents a robot pose at which a measurement was taken.
The vertices are connected with edges that carry the relative transformation $\relpose$ between the two poses.
The edges are either established between subsequent measurements, in which case $\relpose$ is estimated using odometry, or they are established due to place recognition, in which case $\pose{R_{k-x}}{R_k}$ comes from a relative pose estimation algorithm.
The pose graph and our representation can be constructed incrementally and on-line.
The relative transformation between non-adjacent poses can be estimated by integrating relative transformations along a path connecting the two corresponding vertices.
Considering odometry drift, this estimate becomes less accurate as the path length increases.
Without global consistency, estimates integrated along different paths can differ \cite{Bosse03icra,Es15crv}.
Thus, for our purposes, we use the estimate resulting from integrating along the shortest path.

\subsection{Frontier Consolidation}\label{sec:consolidation}

\begin{figure}
\centering
\vspace{-8mm}
\subfloat[Before loop closure]{\includegraphics[width=0.48\linewidth]{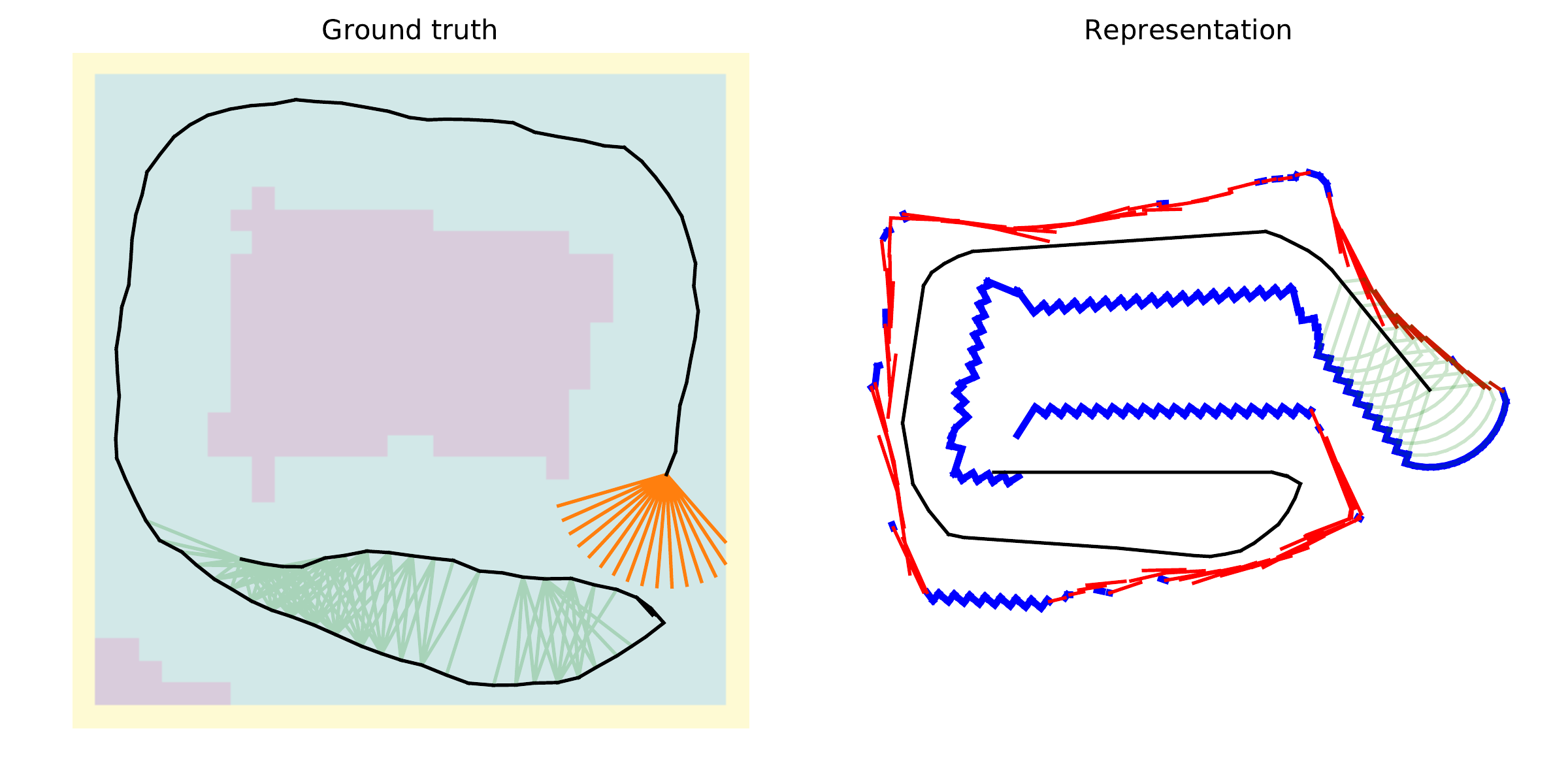}}
\hfil
\subfloat[After loop closure]{\includegraphics[width=0.48\linewidth]{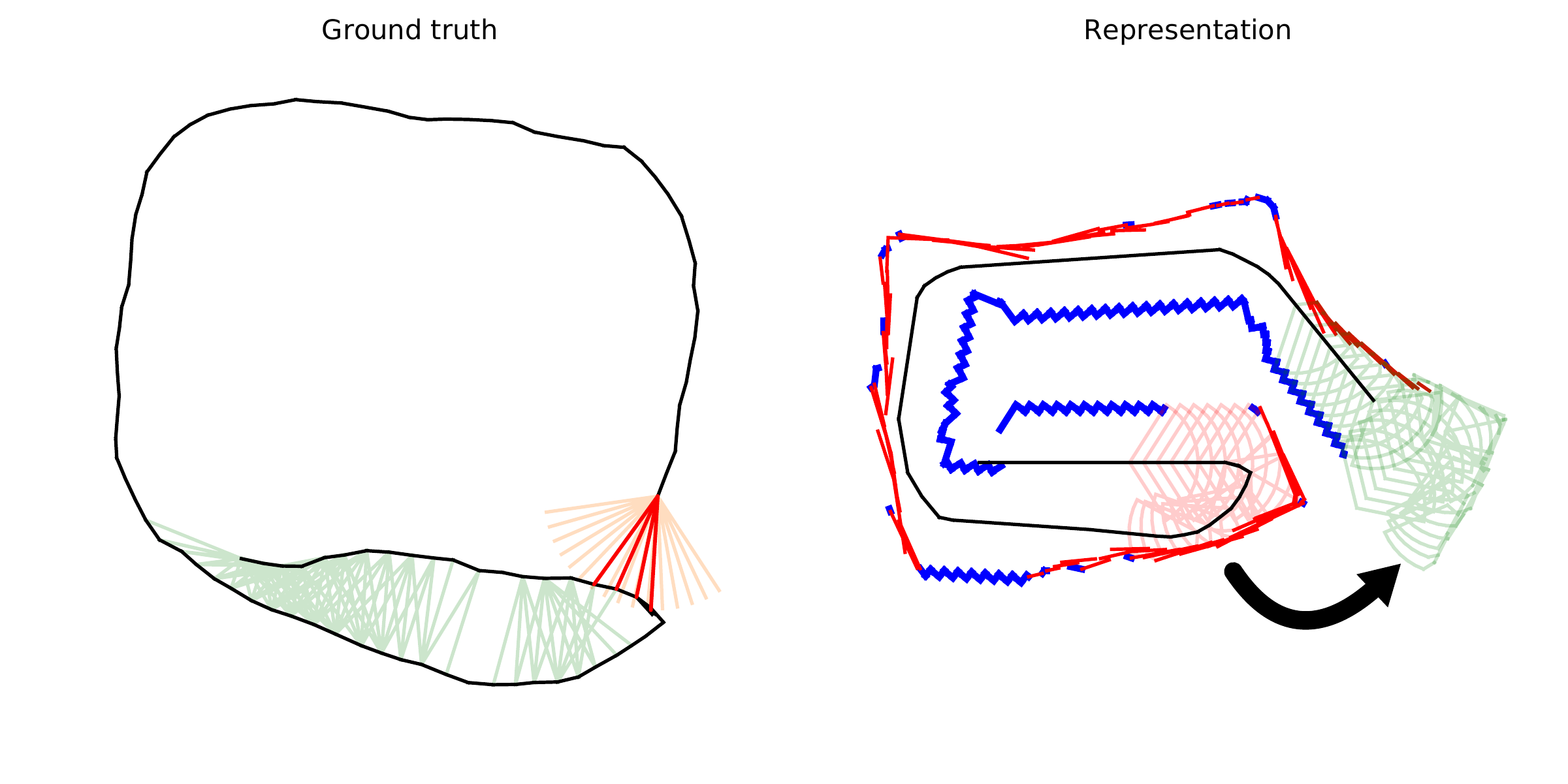}}
\caption{
Frontier consolidation prompted by place recognition.
See \reffig{fig:repr} for a detailed legend.
(a) Before place recognition, the {\color{PRGreen} consolidation scope} contains most recent poses.
(b) After {\color{red} place recognition (red lines in ground truth)}, {\color{ConsRed} local volumes across the place recognition edges (light red)} are added to the consolidation scope, temporarily transformed into the local frame of the current pose (arrow, light green volumes).
This affects {\color{blue} frontiers (blue)}: They frontiers above the {\color{ConsRed}red volumes} and the {\color{blue} \qmarks{range arc}} of the current pose are both resolved.}
\label{fig:consolidation}
\vspace{-5mm}
\end{figure}

When adding new depth measurements, we need to update the local polygons.
New measurements can turn unknown space into known free space, and the corresponding frontier boundaries need to be converted into free boundaries.
Similarly, frontiers might need to be re-assessed after place recognition.
This process, which we call {\it frontier consolidation}, is illustrated in~\reffig{fig:consolidation}.
For every new depth measurement $\vfovk$, a {\it consolidation scope} comprising the local volumes of nearby poses is established using Dijkstra's algorithm.
All poses within a distance of $\consrad$ are added to that scope, where the norm of the estimated translation $t_{A,B}$ between two adjacent poses is used as the weight of the edge between them.
Consequently, $\consrad$ reflects the distance over which the pose estimation (odometry and relative pose estimation) has small drift and can be used to consolidate local volumes.
Inside the consolidation scope, all local volumes are then consolidated pairwise among all possible pairs.
To that end, all local volumes are temporarily expressed relative to $\vfovk$ by using the pose resulting from pose integration along the shortest path from $\vfovk$, see \reffig{fig:consolidation}.
Given two local volumes transformed in this way, any frontier edge of one volume that lies inside the other volume is re-labeled as free edge, and vice versa.
If a frontier edge is only partially inside, it is subdivided accordingly.

Consolidation is triggered for every new depth measurement.
In our experiments, we assume that previously visited places are recognized at the same time as a vertex is inserted into the polygon.
Otherwise, polygon consolidations would also need to be triggered as new edges are inserted into the pose graph due to place recognition.

So while our approach does not rely on odometry accuracy, it relies on good place recognition performance.
An increasing rate of false negatives could potentially be dealt with by methods that expand existing matches, or by increasing the volume in which polygons are consolidated.
An increasing rate of false positives (perceptual aliasing) would be harder to deal with.
However, false positives have also been shown to be catastrophic for optimization-based approaches \cite{Sunderhauf12iros} -- remedies to false positives are equally applicable to both approaches.

\subsection{Extension to 3D}
While for simplicity, we focus on two dimensions in this paper, we believe that the presented approach is readily applicable in 3D.
The main challenge here is the 3D representation of local volumes, and the intersection of those local volumes, where volume boundaries become surfaces, rather than edges.
One could use local occupancy grids, or local coarse meshes \cite{Teixeira16iros, Greene17iccv} for memory efficiency.

\section{Application to Frontier-Based Exploration}

To demonstrate how our map representation can be fitted to exploration approaches that previously used metric representations, we adapt the state-of-the-art method proposed in~\cite{Cieslewski17iros} to use our solution.
Exploration in~\cite{Cieslewski17iros} is performed using a state machine with the following three states:
\begin{enumerate}
\item {\bf Reactive:} While frontiers are present in the current \ac{FOV}, navigate towards the frontier that incurs the least change to the current velocity.
\item {\bf Deliberate:} If no frontiers are left in the \ac{FOV}, but frontiers are left globally, plan a path to the closest frontier.
\item If no frontiers are left overall, exploration is considered {\bf complete}.
\end{enumerate}
Since the {\bf reactive} state is based on the current field of view, it is trivial to adapt to our representation.
In the original implementation, information was extracted from OctoMap~\cite{Wurm10icraw} updates to determine frontier voxels in the current FOV.
With our representation, frontier candidates are simply selected from the edges of the current local polygon.
The {\bf deliberate} state cannot be adapted directly.
In the original implementation, a path to the closest frontier was found using a Dijkstra search across adjacent free space voxels.
With our representation, adjacency of free space is only meaningful among polygons that are also close to each other in the pose graph.
Hence, instead of looking for the closest frontier in 2D space, we instead look for the closest vertex in the pose graph that has a local volume with unconsolidated frontiers.
We then let the robot navigate to that vertex using the teach and repeat navigation method proposed in~\cite{Es15crv}, as it does not require global consistency of the map.
Once the robot arrives at that vertex, it switches back to the reactive state.
{\bf Completion} in the original method is assumed once there are no frontier voxels left.
In our approach, this corresponds to none of the polygons having frontier edges left.

\section{Experiments}

Using~\cite{Cieslewski17iros} as exploration algorithm, we compare our representation to a metric grid-based representation, with and without loop closure capability, in simulated 2D environments under varying conditions.
The representations are compared in different environments, with simulated odometry drift of varying intensity and with simulated place recognition of varying recall.

\subsection{Experimental Setup}

\begin{figure}
  \centering
  \vspace{-5mm}
  \includegraphics[width=0.3\columnwidth]{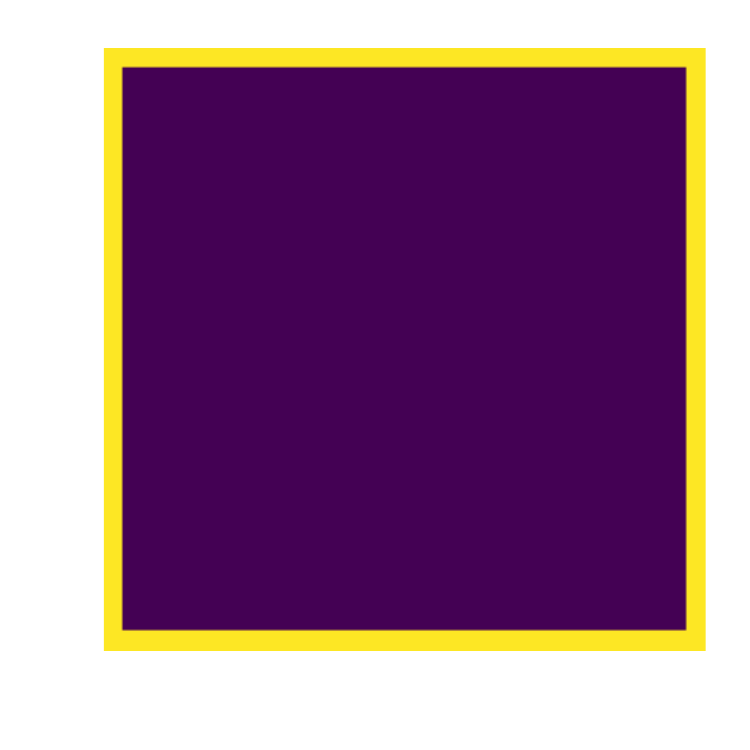}
  \includegraphics[width=0.3\columnwidth]{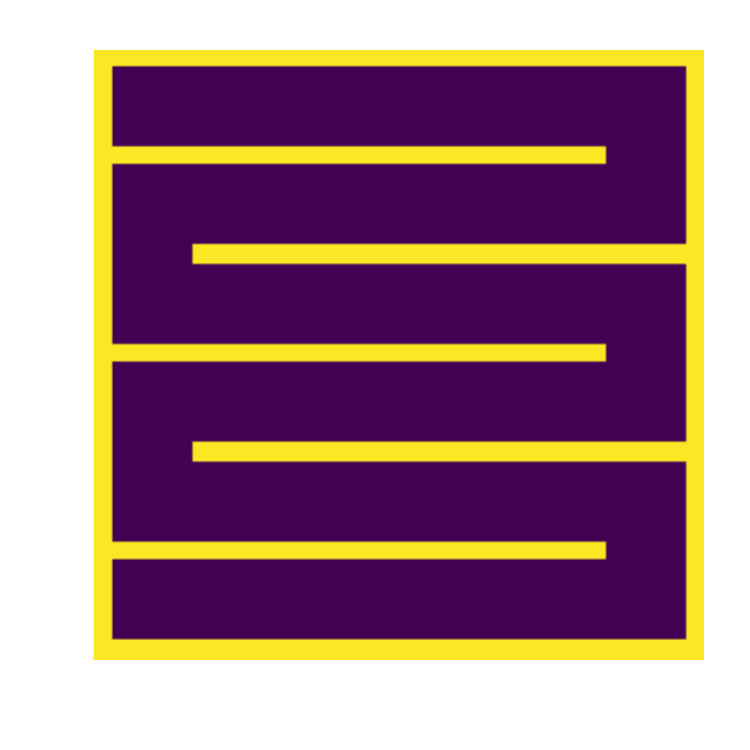}
  \includegraphics[width=0.3\columnwidth]{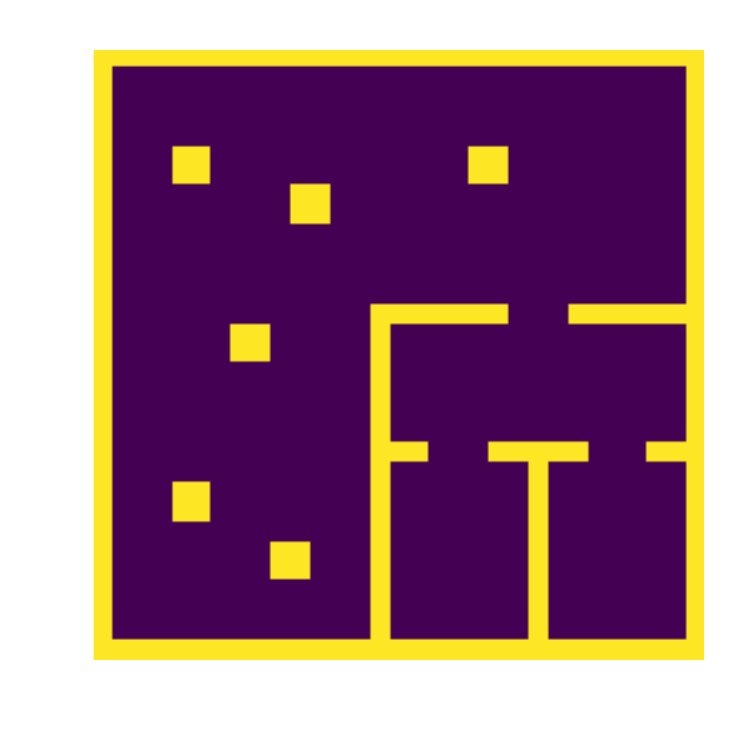}
  \vspace{-5mm}
  \caption{The simulated environments: \qmarks{Open}, \qmarks{Maze} and \qmarks{Forest house}.  
  All have a size of $30 \times 30$ meters.}
  \label{fig:maps}
  \vspace{-5mm}
\end{figure}

We simulate a robot with a forward-looking depth sensor operating in either of the $30m \times 30m$ environments shown in~\reffig{fig:maps}.
The depth sensor has a \ac{FOV} of $115^{\circ}$ in horizontal direction and a range $\dfov$ of $5m$, similar to the sensor used in \cite{Cieslewski17iros}.
The simulation is performed in time steps $k \in \mathbb{N}$, for each of which the robot's true pose $\pose{W}{R_k}$ and pose estimate $\posest{W}{R_k}$ are updated as follows:
\begin{align}
\pose{W}{R_k} &= \pose{W}{R_{k-1}}\posecomm \noise{\relpose}, \\
\posest{W}{R_k} &= \posest{W}{R_{k-1}}\posecomm,
\end{align}
where $\posecomm$ is the relative pose command output provided by the exploration planner.
In the true pose update, a pose increment $\noise{\relpose}$ is applied to $\posecomm$.
As it is the robot that executes the desired pose increment, it believes it has executed $\posecomm$, and so noise needs to be applied to the true pose update, rather than the estimate.
The translation of $\noise{\relpose}$ has coefficients sampled from a zero-mean Gaussian $\mathcal{N}(0, f(\sigma_{\text{pos}}, \posecomm))$ and its rotation angle is sampled from $\mathcal{N}(0, f(\sigma_{\text{rot}}, \posecomm))$.
Here, $f$ is a function that ensures that the variance of the noise is proportional with the distance travelled:
\begin{equation}
    f(\sigma, \posecomm) = \norm{\posiest{R_{k-1}}{R_k}} \cdot \sigma^2 .
\end{equation}
We furthermore simulate place recognition.
Whenever the current pose $\pose{W}{R_k}$ falls within a threshold distance $\lcd$ to a previous pose $\pose{W}{R_l}$, the identifier of that pose, $l$, as well as $\pose{R_l}{R_k}$ are provided to the simulated robot.
To prevent self-matches, all poses within $1.5 \cdot \lcd$ of pose graph traversal are excluded from place recognition.
Also, place recognition is only provided if $\pose{W}{R_l}$ is in line of sight from $\pose{W}{R_k}$, to prevent place recognition across occlusions.

\subsection{Baseline Implementation}

As a baseline, we implement a simplified version of the grid representation \cite{Wurm10icraw} using a regular grid instead of octrees, as we do not take computational performance into account.
According to \cite{Wurm10icraw, Moravec85icra}, the occupation probability $P(n|z_{1:k})$ of each cell is updated using the most recent measurement $z_k$ with
\begin{equation}
L(n|z_{1:k}) = L(n|z_{1:k-1}) + L(n|z_k), \quad L(n) = \lim_{x \to P(n)} \log (\frac{x}{1 - x}).
\label{eqn:gridup}
\end{equation}
We assume a precise depth sensor and let $P(n|z_k) = 1$ if a ray hits an obstacle inside $n$ at time $k$, $P(n|z_k) = 0$ if cell $n$ is fully contained inside the polygon spanned by the sensor rays as defined in \refsec{sec:construction}, and $P(n|z_k) = 0.5$ otherwise.
Consequently, $P(n)$ can only assume values $\in \{0, 0.5, 1\}$, which we accordingly label as {\it free}, {\it unknown} and {\it occupied}.
For cases where \eqref{eqn:gridup} is undefined, later measurements override earlier measurements.
This baseline is implemented in two ways:
a naive one (\qmarks{{\it grid}}), which uses the state estimate $\posest{W}{R_k}$ as-is, and one with loop closure capability (\qmarks{{\it grid-lc}}).
The latter exploits place recognition events to optimize its pose-graph according to all relative pose estimates present in the pose graph using g2o \cite{Kuemmerle11icra}.
Every place recognition triggers an optimization after which all updates according to \eqref{eqn:gridup} are executed anew with the optimized pose estimates $\posest{W}{R_k}^\star$.
We use a voxel length of 1m for all maps.

\subsection{Evaluation Metrics}

To quantify the performance of exploration, we measure the distance traveled until the robot {\it believes} full coverage has been achieved, $d_\text{max}$, and the expected distance to be traveled to discover an arbitrary location in free space, $d_\text{exp}$.
$d_\text{exp}$ reflects the general speed of exploration.
It is not severely affected if the robot takes a very long time at the end to navigate to the last couple of frontiers.
Depending on the application, one metric is more important than the other.
For more insight into the progress of exploration, we track the ratio between the volume of known free space $\kfs$ and the full free space, $\ffs$, $|\kfs|/|\ffs|$, which we call {\it coverage ratio}.
It is initially $0$ and reaches $1$ when full coverage is achieved.
Note that $\kfs$ refers to the ground truth volume covered by the robot, not the estimate of that space by the robot.
$\ffs$ is represented using a grid -- in our simulations we use this grid to define the environment in the first place, see \reffig{fig:maps}.
Unlike in the baseline grid representation, occupied cells are known beforehand, and only free cells $n \in \ffs$ are labeled as {\it known free} or {\it unknown free}.
Known free space $\kfs$ is approximated as the set of cells labeled {\it known free}.
An unknown free cell is relabeled {\it known free} as soon as any sensor ray intersects it and remains {\it known free} until the end of the simulation.
Given this approximation of $\kfs$, the coverage ratio is calculated  by dividing the count of cells in $\kfs$ by the count of cells in $\ffs$.
Since full coverage will not be reached in all experiments, even if the robot believes this to be the case, we also report the final coverage ratio.
In these cases, $d_\text{exp}$ is undefined.
\begin{figure}
  \centering
  \vspace{-10mm}
  \subfloat[Open]{\includegraphics[trim=0 0 41 0,clip,width=0.33\columnwidth]{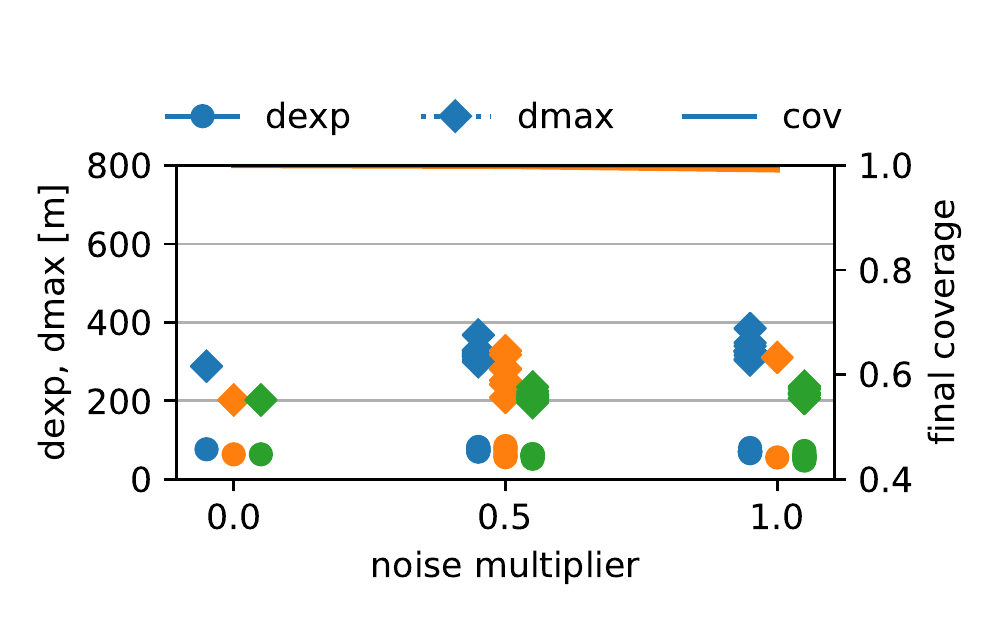}}
  \subfloat[Maze]{\includegraphics[trim=45 0 41 40,clip,width=0.27\columnwidth]{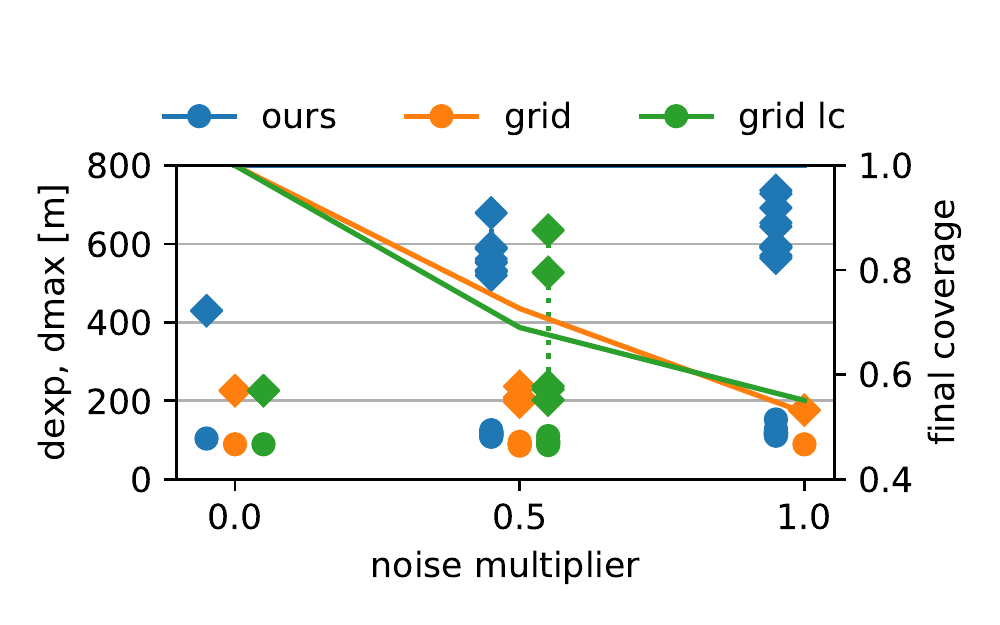}}
  \subfloat[Forest house]{\includegraphics[trim=45 0 0 0,clip,width=0.326\columnwidth]{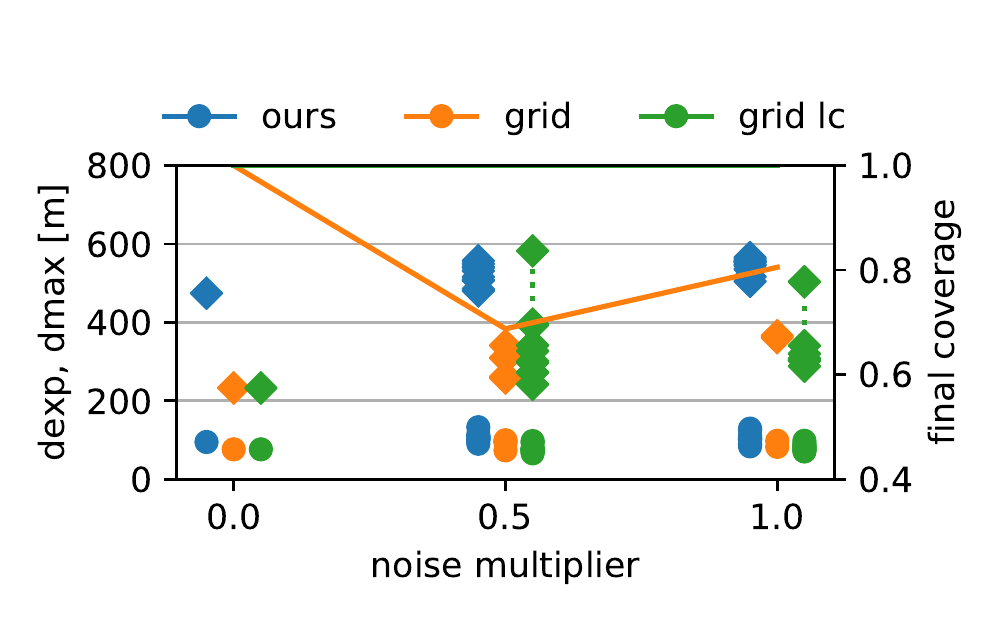}}
  \caption{Expected distance until discovery $\dexp$ and distance until termination $\dmax$ versus final coverage for the evaluated approaches, with different noise intensity, on different maps.
  For fair comparison, $\dexp$ and $\dmax$ are omitted for samples where the final coverage is below $1$.
  {Ten samples are collected for every setting.}}
  \label{fig:dmax}
  \vspace{-5mm}
\end{figure}

\subsection{Experiments}

We perform experiments with the following parameter combinations:
simulation in either of the environments shown in \reffig{fig:maps};
pose estimate noise simulated with
\begin{equation}
\sigma_{\text{pos}} = \alpha \cdot 0.1m, \quad \sigma_{\text{rot}} = \alpha \cdot 5^\circ,
\end{equation}
with the noise multiplier $\alpha \in \{0, 0.5, 1\}$;
and place recognition radius $d_\text{pr}$ of $0.5, 1, 1.5$ or $2$ times the sensor range $\dfov$.
For every parameter setting, ten different runs have been performed (due to the stochastic nature of the simulated noise).
The place recognition radius is varied because a larger radius should result in the robot having to travel less in order to consolidate frontiers.
It is limited to two times the sensor range as beyond that, polygon intersection between $\vfovk$ and $\vfovof{l}$ is impossible.

\section{Results}
\begin{figure}
  \centering
  \vspace{-10mm}
  \subfloat[Open]{\includegraphics[width=0.34\columnwidth]{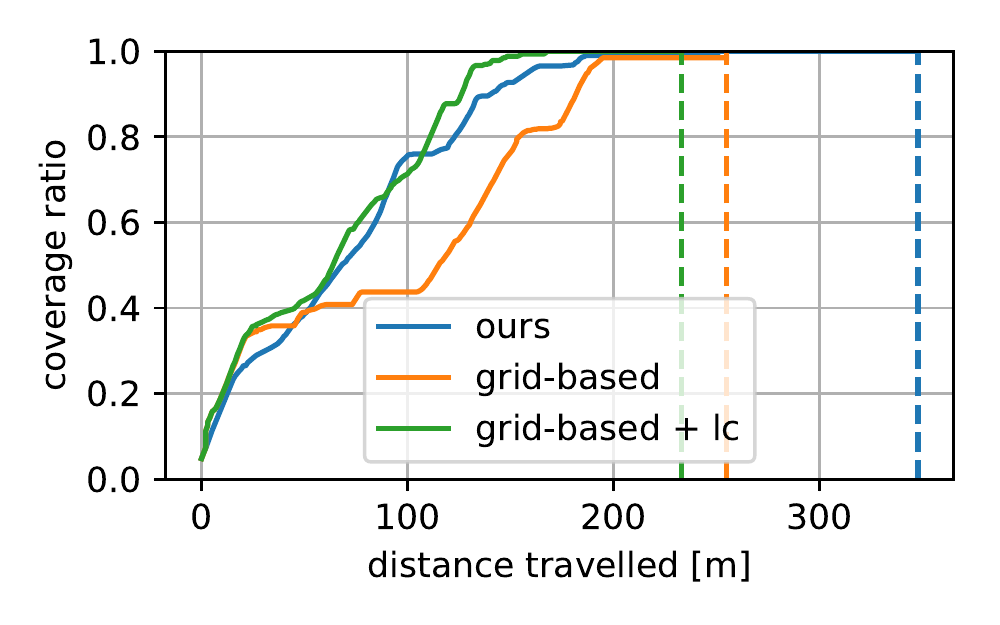}}
  \subfloat[Maze]{\includegraphics[trim=45 0 0 0,clip,width=0.285\columnwidth]{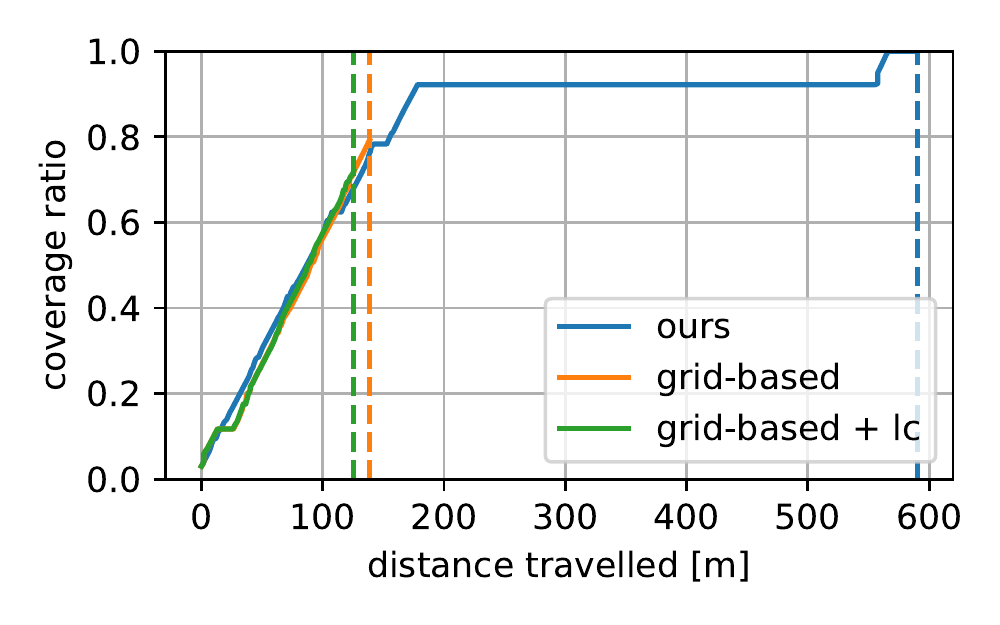}}
  \subfloat[Forest house]{\includegraphics[trim=45 0 0 0,clip,width=0.285\columnwidth]{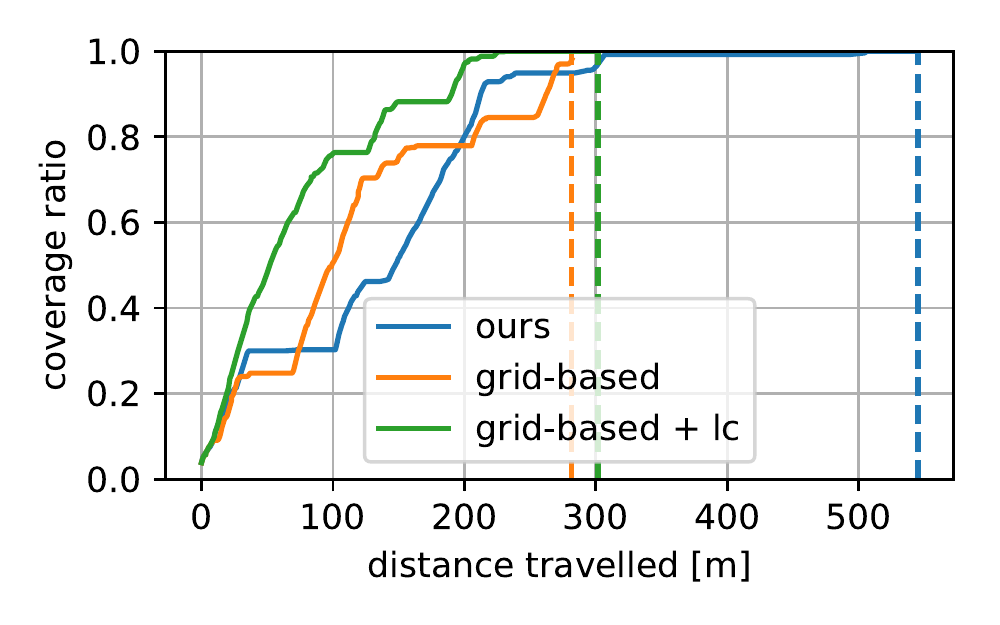}}
  \caption{True coverage as a function of distance travelled for individual runs at noise multiplier $\alpha = 1$.
  Vertical dashed lines indicate the distance at which the robot believes coverage to be complete according to its own representation.}
  \label{fig:covd}
  \vspace{-5mm}
\end{figure}

\reffig{fig:dmax} shows the performance of the evaluated methods as pose estimation noise is increased.
As we can see, only the proposed representation always reaches full coverage in all environments.
The other approaches perform particularly poorly in the maze environment, where the robot has to travel long distances in close proximity without the ability to close loops, see \reffig{fig:eyecat}.
We can also see that this comes at a cost of longer distances covered until the robot believes coverage is achieved.
$\dmax$ is { between $1.5$ and $3$ times} larger for our approach, $\dexp${, however, only up to $1.5$ larger}.
This happens because the proposed approach needs to consolidate frontiers of all spatially close polygons through place recognition, while the grid-based representation is able to consolidate frontiers from measurements even if the relative pose established between them comes from an estimate integrated over a very long path.
A typical evolution of the coverage ratios for one sample of each method and noise multiplier $\alpha = 1$ is shown in \reffig{fig:covd}.

\begin{figure}
	\centering
	\subfloat[Open]{\includegraphics[width=0.4\columnwidth]{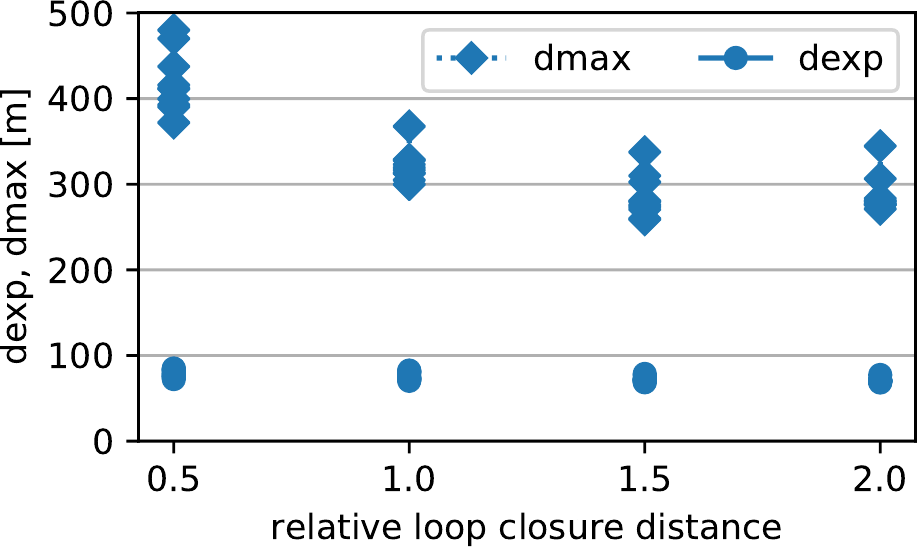}}
	\subfloat[Maze]{\includegraphics[width=0.4\columnwidth]{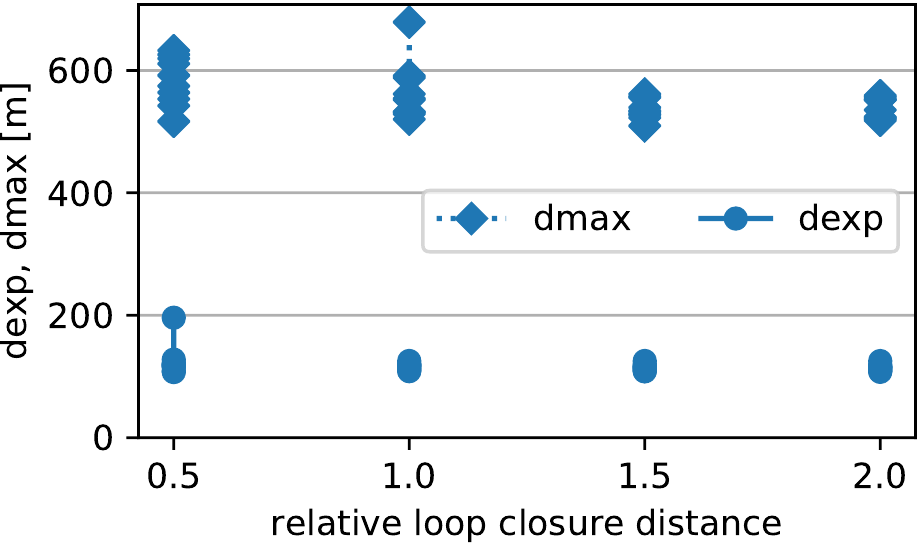}}
  \caption{$\dexp$ and $\dmax$ when using our representation in the {\it open} and {\it maze} environments with noise multiplier $\alpha = 0.5$, for different relative loop closure distances $\beta$.
  Place recognition is provided by the simulator for any two poses when the distance between them is below $\beta$ times the depth sensor range.}
	\label{fig:rlcd}
	\vspace{-5mm}
\end{figure}

\reffig{fig:rlcd} shows how our method is affected by the loop closure distance $\lcd$.
As can be seen, a larger distance at which places can be recognized leads to faster exploration times in the open environment.
This makes sense, as a larger loop closure distance leads to generally larger frontier consolidation scopes, allowing frontiers to be removed faster.
Exploration speed in the maze environment, however, is not significantly affected, as there are generally not many place recognition events happening in that environment.

\section{Validation in the Real World}

\begin{figure}
\centering

\includegraphics[width=0.39\linewidth]{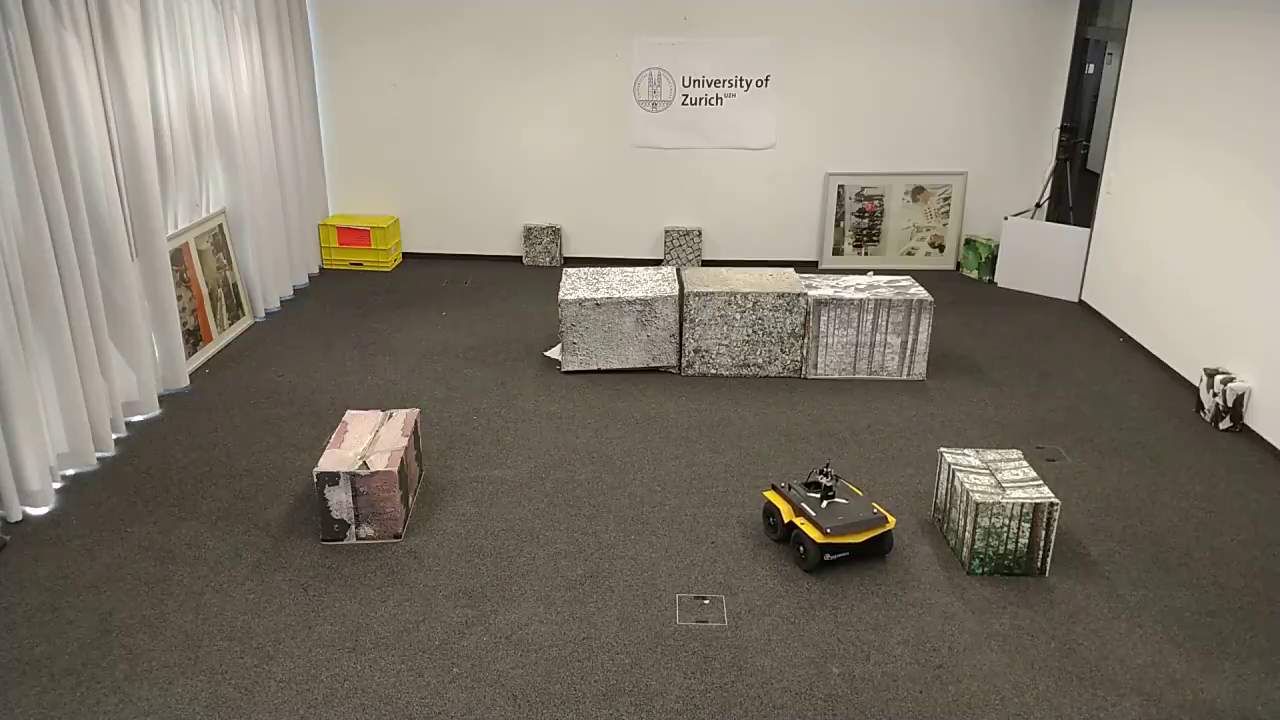} ~
\includegraphics[width=0.29\linewidth]{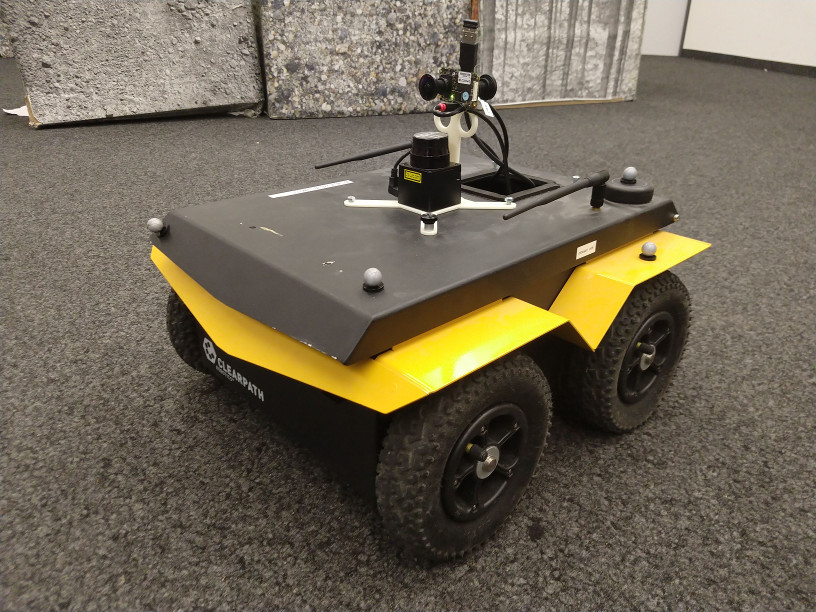} ~
\includegraphics[width=0.27\linewidth]{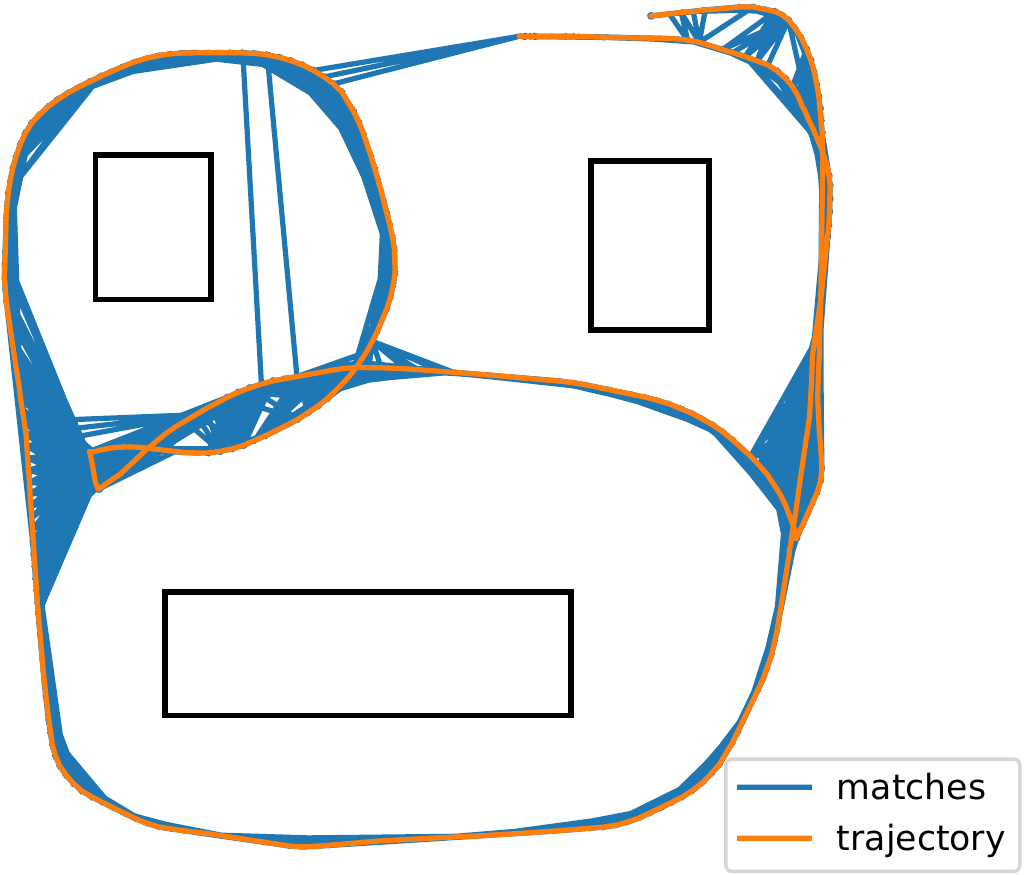}

\caption{(a) Real environment in which we have validated our approach.
(b) Close-up of our experimental platform.
(c) NetVLAD matches on a trajectory circumnavigating all obstacles.
Obstacle locations are approximate.}
\label{fig:real_env}
\vspace{-5mm}
\end{figure}

We validate that our approach works in the real world with the experimental setup shown in \reffig{fig:real_env}.
The platform is a ClearPath {\it Jackal} equipped with a Hokuyo laser scanner as depth sensor and two fisheye cameras that provide a $360^\circ$ view of the environment.
Ground truth is obtained with a motion tracking system using reflective IR markers.
For the state estimate, we use the wheel odometry provided by the Jackal robot.
We intentionally do not use the best available state estimate, to demonstrate robustness to drift.

The most important assumption that we have made in our simulations is that sufficient place recognition and relative pose estimation can be provided.
To this end, we use visual place recognition from a panoramic image stitched from the two fisheye cameras. %
The vertical field of view is restricted to prevent place recognition from structure that is visible from everywhere in the room.
Using a $360^\circ$ view allows place recognition independent of the orientation of the robot between the two matched places.
First, the CNN full image desciptor NetVLAD \cite{Arandjelovic16cvpr} is used to quickly determine the visually most similar previously recorded image.
\reffig{fig:real_env} (c) shows NetVLAD matches in a trajectory where the robot drives once around each obstacle.
As can be seen, all place recognitions occur locally, with only two matches across a longer distance, but still within line of sight.
Given a NetVLAD match, we match ORB features \cite{Rublee11iccv} and use P3P \cite{Gao03pami} and RANSAC \cite{Fischler81cacm} for geometric verification and relative pose estimation.
For P3P, the 3D locations of features in each frame are triangulated using KLT tracks \cite{Lucas81ijcai} of those features in subsequent frames.
The only component that we have not implemented, as it is out of the scope of this project, is teach-and-repeat \cite{Es15crv}.
To simulate teach-and-repeat, we instead use the motion tracking system to let the robot backtrack its trajectory.
This, and ground truth for evaluation, are the only things for which the motion tracking system is used.

\begin{figure}
\centering
\vspace{-7mm}
\subfloat[]{\includegraphics[width=0.53\linewidth]{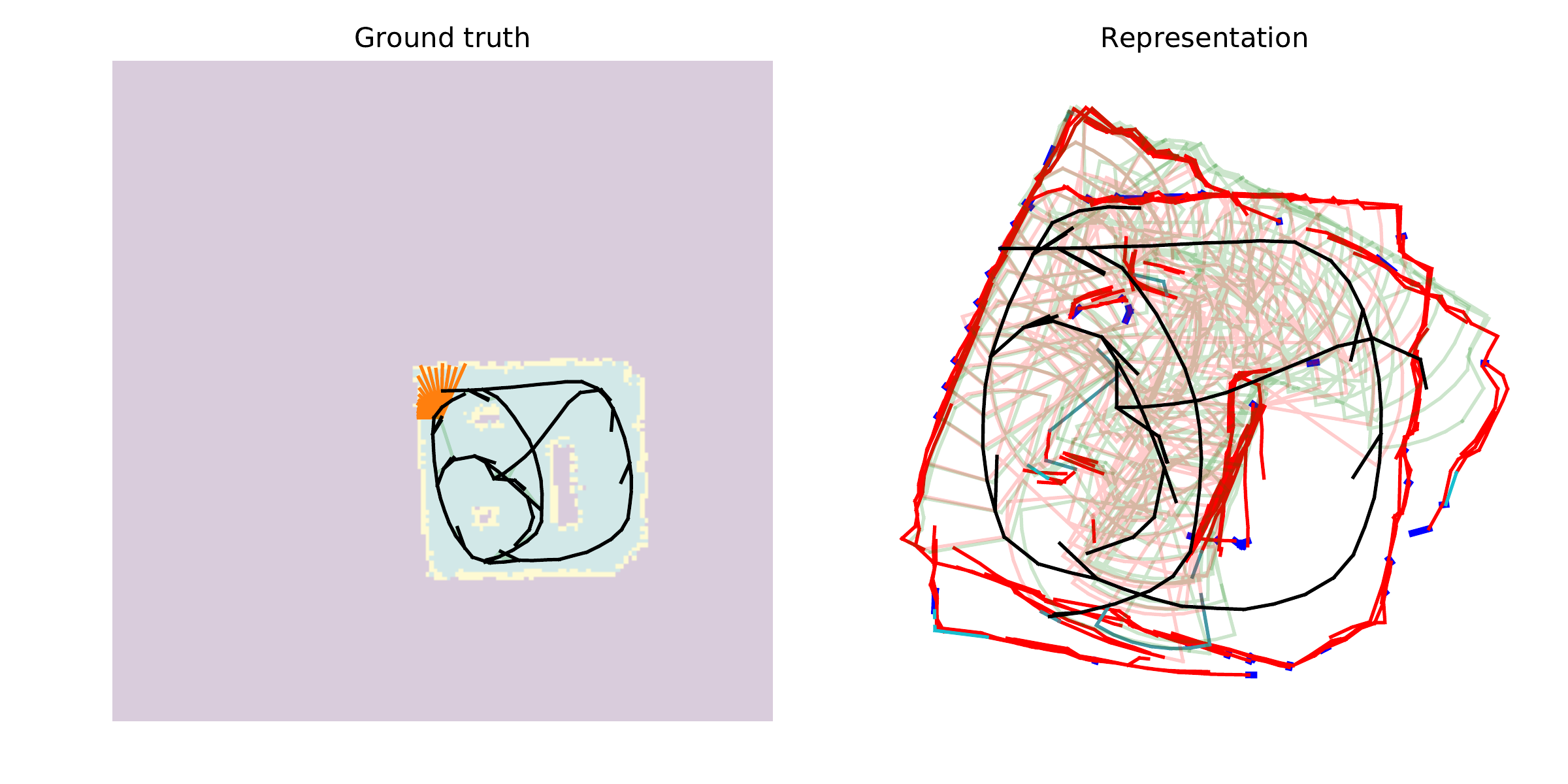}}
\subfloat[]{\includegraphics[width=0.38\linewidth]{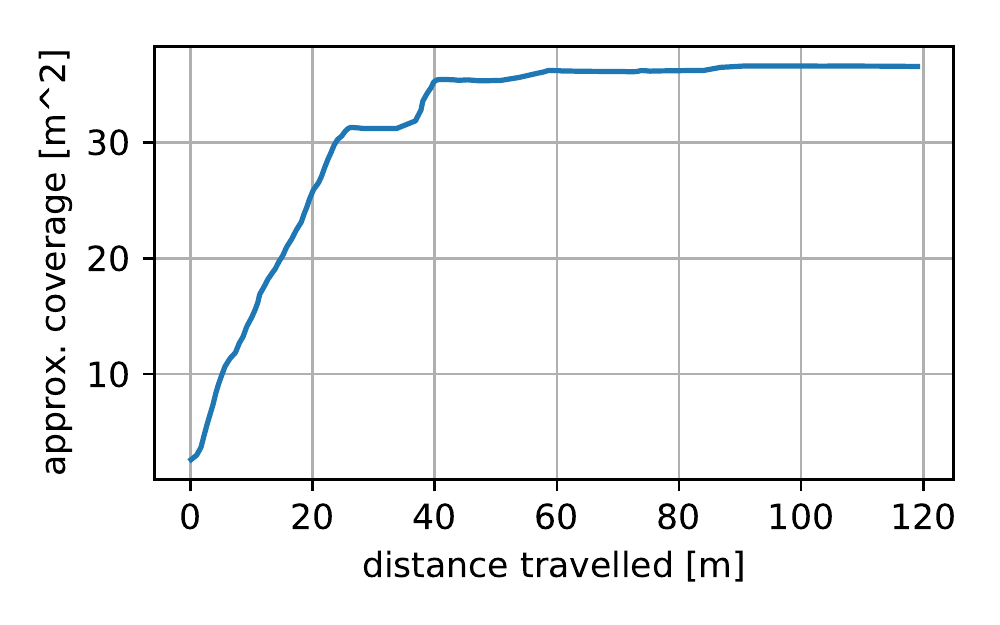}}
\caption{(a) Ground truth and representation once exploration is complete.
(b) Corresponding coverage over time, obtained from the ground truth grid.
}
\label{fig:real_res}
\vspace{-5mm}
\end{figure}

\reffig{fig:real_res} (a) shows the final state of exploration in our experiment.
As we can see, there is drift in the estimated trajectory, yet our approach still manages to fully cover the environment.
Note that there are parts of the environment where trajectories overlap.
These are locations where our visual relative pose estimator fails to obtain enough feature matches.
However, as long as relative pose estimations are obtained within the distance of the consolidation scope (see polygons shown in \reffig{fig:real_res} (a)), this does not pose a significant problem.
\reffig{fig:real_res} (b) shows the corresponding coverage over the travelled distance.
The behaviour is consistent with the results obtained in simulation, wherein the robot first quickly covers most of the map, after which it does a lot of backtracking to seek out frontiers that remain in the map.

\section{Conclusion}

In this paper, we present a map representation that allows exploration in spite of large drift in the pose estimate.
We show that global consistency in the pose graph is not required to determine if exploration is complete.
This alleviates the need for map optimization, which is particularly interesting for multi-robot exploration.
In addition, the proposed method can be adapted to algorithms that currently use different representations.
Using a state-of-the-art exploration algorithm, we compare our representation to a grid-based representation.
In contrast to the latter, and at a cost of longer exploration time, all of the free space can be fully covered with our representation, even with large drift in the state estimate.

\section*{Acknowledgments}
{ This work was supported by the National Centre of Competence in Research (NCCR) Robotics through the Swiss National Science Foundation and the SNSF-ERC Starting Grant.}

\bibliographystyle{spmpsci}
\bibliography{all}

\end{document}